\documentclass[10pt,twocolumn,letterpaper]{article}

\usepackage{iccv}
\usepackage{times}
\usepackage{epsfig}
\usepackage{graphicx}
\usepackage{amsmath}
\usepackage{amssymb}
\usepackage{booktabs}

\usepackage[breaklinks=true,bookmarks=false]{hyperref}
\usepackage{times}
\usepackage{epsfig}
\usepackage{graphicx}
\usepackage{subcaption}
\usepackage{multirow}
\usepackage{amsmath}
\usepackage{color}
\usepackage{amssymb}
\usepackage{bbding}
\usepackage{algorithm}
\usepackage{algpseudocode}

\usepackage[capitalize]{cleveref}
\crefname{section}{Sec.}{Secs.}
\Crefname{section}{Section}{Sections}
\Crefname{table}{Table}{Tables}
\crefname{table}{Tab.}{Tabs.}

\iccvfinalcopy 

\def\iccvPaperID{6004} 

\ificcvfinal\fi

\begin{document}

\title{Learned Image Reasoning Prior Penetrates Deep Unfolding Network for Panchromatic and Multi-Spectral  Image Fusion}

\author{Man Zhou$^{1}$\thanks{Co-first authors contributed equally, $^\dagger$ corresponding author. We also gratefully acknowledge the support of MindSpore, CANN, and Ascend AI Processor used for this research.},
Jie Huang$^{2}$\footnotemark[1], Naishan Zheng$^{2}$, Chongyi Li$^{3\dagger}$\\
$^{1}$S-Lab, Nanyang Technological University,
Singapore\\
$^{2}$University of Science and Technology of China, China\\
$^{3}$Nankai University, China\\
}

\maketitle
\ificcvfinal\thispagestyle{empty}\fi

\begin{abstract}


The success of deep neural networks for pan-sharpening is commonly in a form of black box, lacking transparency and interpretability.
To alleviate this issue, we propose a novel model-driven deep unfolding framework with image reasoning prior tailored for the pan-sharpening task. 
Different from existing unfolding solutions that deliver the proximal operator networks as the uncertain and vague priors, our framework is motivated by the content reasoning ability of masked autoencoders (MAE) with insightful designs.
Specifically, the pre-trained MAE with spatial masking strategy, acting as intrinsic reasoning prior, is embedded into unfolding architecture.
Meanwhile, the pre-trained MAE with spatial-spectral masking strategy is treated as the regularization term within loss function to constrain the spatial-spectral consistency.
Such designs penetrate the image reasoning prior into deep unfolding networks while improving its interpretability and representation capability.
The uniqueness of our framework is that the holistic learning process is explicitly integrated with the inherent physical mechanism underlying the pan-sharpening task. 
Extensive experiments on multiple satellite datasets  demonstrate the superiority of our method over the existing state-of-the-art approaches.  
Code will be released at \url{https://manman1995.github.io/}.     
   
\end{abstract}

\section{Introduction}
\label{sec:intro}

Pan-sharpening, a texture-rich panchromatic image-guided multi-spectral image super-resolution task, is to reason the unknown content at the pre-defined pixel positions according to the context of low-resolution (LR) multi-spectral (MS) image and high-resolution (HR) panchromatic (PAN) image.
%
Owing to the physical constraints, satellites usually adopt both MS and PAN sensors to observe scenes, providing the MS images with high spectral but limited spatial resolution and the PAN images with high spatial but low spectral resolution. 
To obtain  observation with both high spectral and high spatial resolutions, the pan-sharpening technique has drawn increasing attention in both image processing and remote sensing communities. 

\begin{figure}[!t]
	\centering
	\includegraphics[width=0.4\textwidth]{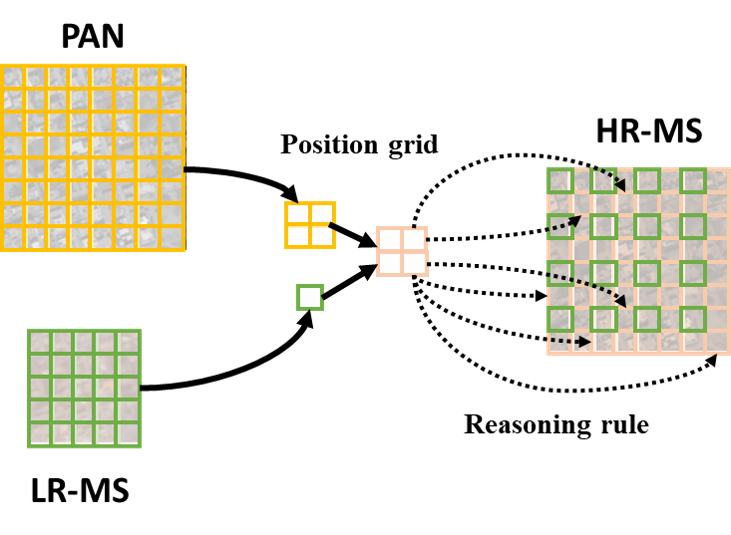}
 \vspace{-5mm}
	\caption{\textbf{Motivation.} Pan-sharpening is to reason the unknown content at the pre-defined pixel positions
according to the context of low-resolution multi-spectral
image and high-resolution panchromatic image.}
	\label{fabs}
	\vspace{-3mm}
\end{figure}

Many research efforts have been devoted to solving the pan-sharpening problem, which can be categorized into two groups: traditional optimization methods and deep learning-based methods. 
Since an infinite number of HR-MS images can be downsampled to produce the same LR-MS image, reasoning the HR-MS images from the LR counterparts is highly ill-posed. 
To solve the ill-posedness, various natural images priors as regularization terms have been developed  in traditional optimization methods, \eg, low-rank prior \cite{ye2019pan} and sparse image priors \cite{zhu2012sparse}. 
However, these priors are not easy to be devised.
Moreover, it is challenging to optimize these methods, hampering the practical applications. Besides, due to the hand-crafted designs, their limited representation ability results in unsatisfactory  performance.

The powerful learning capability of deep neural networks ignites renewed interest in this problem.
As a pioneering work, PNN \cite{masi2016pansharpening} employs three-layer convolution neural networks to account for pan-sharpening learning.
Since then, more and more complicated and deeper architectures have been proposed to improve the performance of pan-sharpening \cite{MHNet2019, yang2017pannet, yuan2018multiscale}. 
Despite the remarkable progress, most of them focus on heuristically constructing  network architectures in a black box fashion without considering the underlying rationality of pan-sharpening task, lacking transparency and interpretability.

\begin{figure}[!t]
	\centering
	\includegraphics[width=0.5\textwidth]{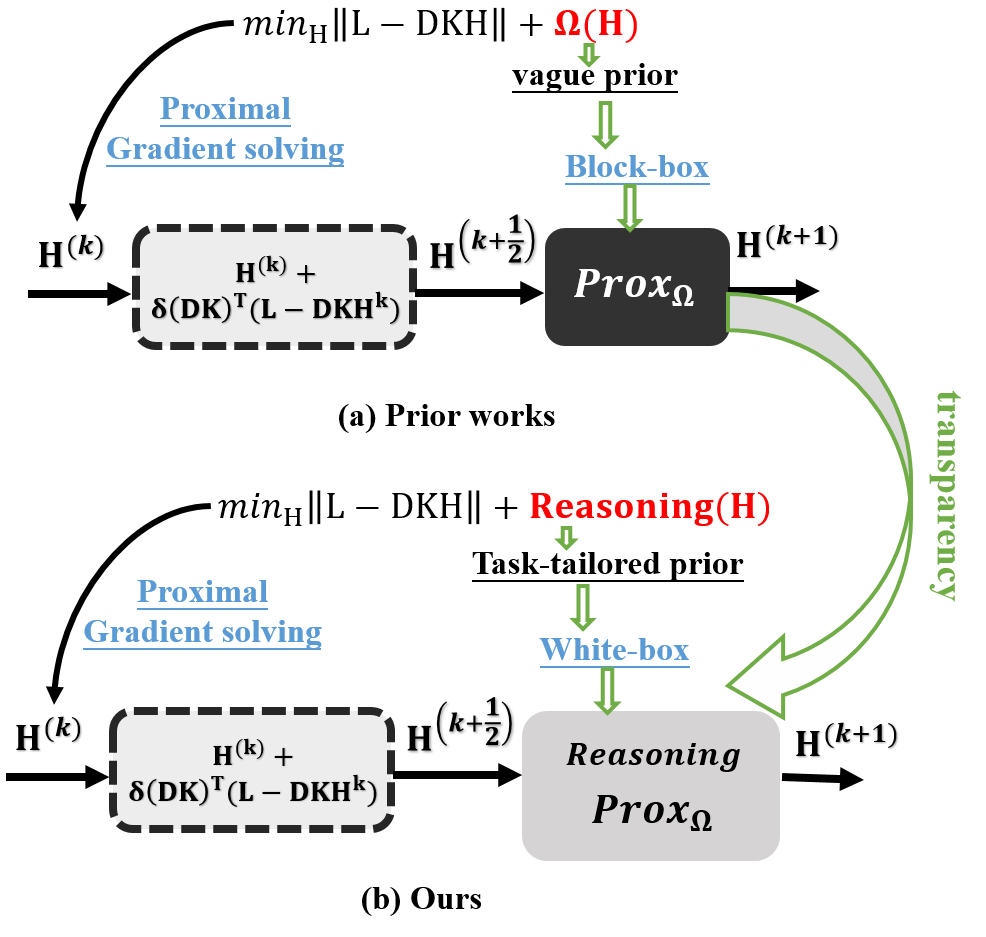}
 \vspace{-5mm}	
 \caption{\textbf{Motivation.} The comparison between prior unfolding frameworks and ours. }
	\label{imagemae1}
	\vspace{-3mm}
\end{figure}

To improve interpretability,  model-driven deep unfolding methods have been proposed. 
Xu \emph{et al.} \cite{Xu_2021_CVPR} propose the first deep unfolding network for pan-sharpening. 
The basic idea behind it is to formulate pan-sharpening as an optimization problem and employ the proximal gradient descent algorithm to solve it. 
The optimization process can be reformulated as:
\begin{align}\label{eq:model}
\vspace{-6pt}
     \hat{\mathbf{H}} = \min\limits_{\mathbf{H}}\frac{1}{2}\left|\left|\mathbf{L}-\mathbf{D}\mathbf{K}\mathbf{H}\right|\right|_2^2 + \lambda\Omega(\mathbf{H}|\mathbf{P}),
     \vspace{-6pt}
\end{align}
where $\lambda$ is a hyper-parameter to weight the first term (data fidelity term accords with the degradation) and the second term (regularization term $\Omega(\cdot)$). $\mathbf{D}$, and $\mathbf{K}$ denote the down-sampling and blur operators, respectively. $\rm L$, $\rm P$, and $\rm H$ respectively represent the LR-MS image, PAN image, and HR-MS image. 
Technically, the proximal gradient descent algorithm can be approximatively expressed as an iterative convergence problem by solving the following iterative function:
\begin{align}
    \rm \mathbf{H}^{(k)}= prox_{\lambda, \Omega, P}(\mathbf{H}^{(k-1)}+\delta(\mathbf{D}\mathbf{K})^{T}(\mathbf{L}-\mathbf{D}\mathbf{K}\mathbf{H}^{(k-1)})), \label{hk1}
\end{align}
where $\rm k$ and $\delta$ denote the iterative step and learning factor.

\noindent
\textbf{Motivation.} 
In terms of the function $\rm prox_{\lambda, \Omega, P}$ that involves the embedded prior term $\Omega(\cdot)$, existing works usually  heuristically employ diverse network architectures in a black box fashion and deliver the proximal operator networks as the uncertain and vague priors, vislized in Figure \ref{imagemae1}.
These methods thus lack clear physical meanings of the prior terms. 
In addition,  most deep unfolding-based methods \cite{Xu_2021_CVPR, yanmemory} cannot extract the pan-sharpening customized prior with clear physical patterns well, which is caused by weak interpretable prior operations. Hence, we argue that the deep unfolding framework with sufficient interpretability has the potential of improving  performance.

\noindent
\textbf{Solution.} 
The first function of pan-sharpening is to reason the unknown information at the pre-defined pixel positions using the context, detailed in Figure \ref{fabs}.
In this work, we propose a novel model-driven deep unfolding framework for pan-sharpening with interpretability.
Inspired by the content reasoning ability of masked autoencoders (MAE) \cite{he2022masked}, our framework endows the holistic learning process of deep unfolding with the explicitly integrated with inherent physical mechanism underlying the pan-sharpening task.
\textit{The key insights of our framework are (1) we embed the pre-trained MAE with spatial masking strategy into the unfolding architecture, acting as intrinsic reasoning prior and (2)
we treat the pre-trained MAE with spatial-spectral masking as the regularization term within loss function to constrain the spatial-spectral consistency.}
Such new designs penetrate the image reasoning prior into  deep unfolding network while improving the interpretability and representation ability.  
Besides, our framework also outperforms existing state-of-the-art approaches on multiple satellite datasets.

Our main contributions are summarized as follows:

\begin{itemize}
 \item We propose to embed the pre-trained MAE into deep unfolding architecture, resorting to its image reasoning ability for pan-sharpening. Such reasoning prior makes the framework transparent and  interpretable.
 
 \item We creatively treat the pre-trained MAE with spatial-spectral masking strategy as regularization term within loss function to constrain the spatial-spectral consistency. The tailored regularization term with intrinsic knowledge of spatial-spectral reasoning empowers the unfolding framework.
 
  \item In contrast to previous works, our framework as the first attempt pushes the frontiers of pan-sharpening towards the designs of the deep prior term. It shows outstanding performance on three satellite datasets,  outperforming the state-of-the-art algorithms.
\end{itemize}

\begin{figure*}[!ht]
\begin{center}
\includegraphics[width=0.9\textwidth]{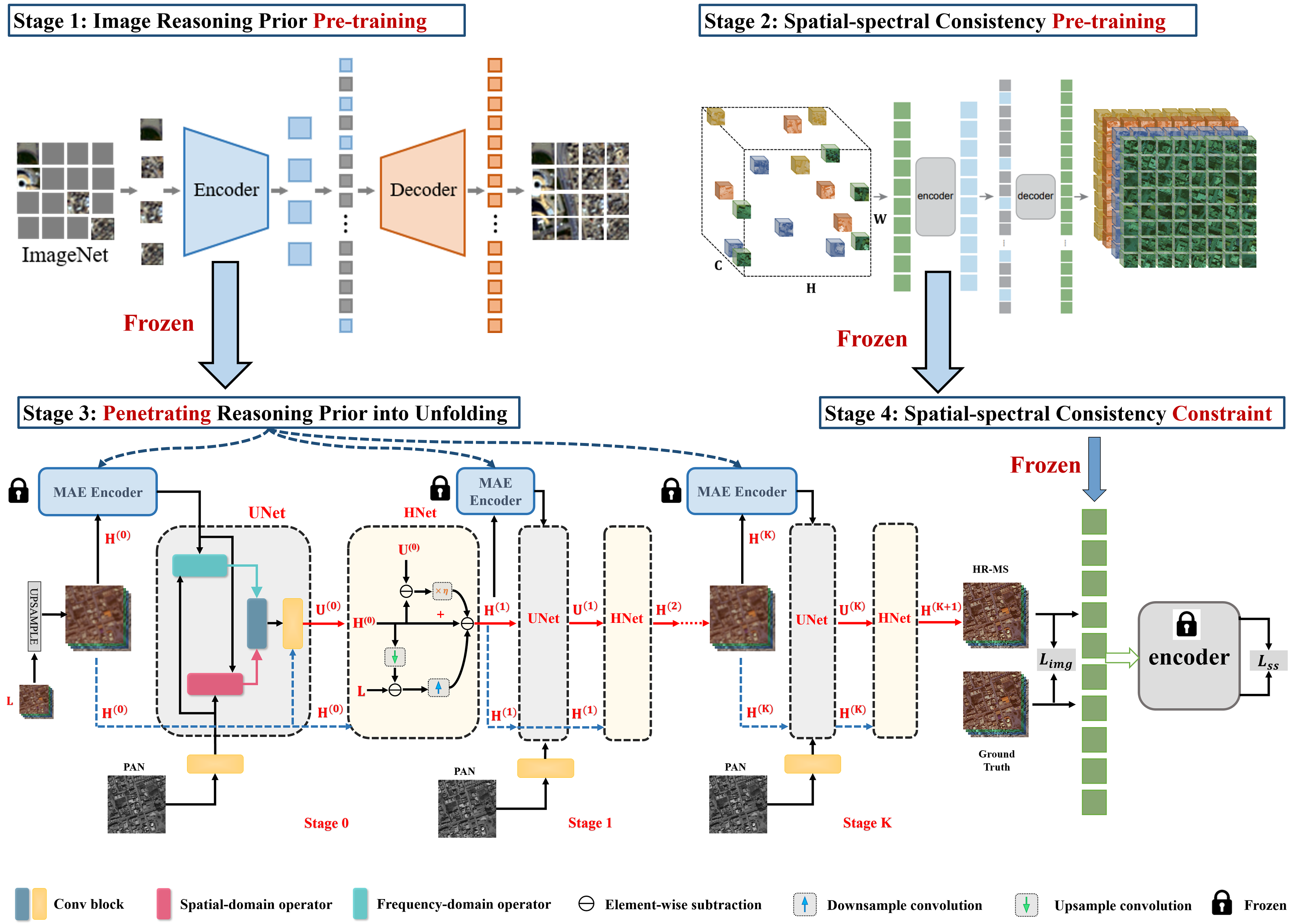}
 \vspace{-3mm}	
\caption{The overall architecture of our proposed method. In detail, LR-MS image is firstly up-sampled and then performs the stage-wise iteration, updating $\rm U$ and $\rm H$ in the overall $\rm K$ stages where the pre-trained MAE acting as reasoning prior penetrates Deep Unfolding process. To promote the spatial-spectral consistency, the pre-trained extension version is employed as the loss constraint. (Best viewed in color.)}
\label{mainfig}
\end{center}
\vspace{-2em}
\end{figure*}

\section{Related Work}

\noindent
\textbf{Traditional Methods.}
Traditional pan-sharpening methods can be roughly divided into three main categories: CS-based methods, MRA-based methods, and VO-based methods. The CS-based approaches separate spatial and spectral information from the LRMS image and replace spatial information with a PAN image. Intensity hue-saturation (IHS) fusion \cite{carper1990use}, the principal component analysis (PCA) methods \cite{kwarteng1989extracting,shah2008efficient}, Brovey transforms \cite{gillespie1987color}, and  Gram-Schmidt (GS) orthogonalization method \cite{GS} are CS-based approaches. These CS-based approaches are fast since LR-MS images simply need spectral treatment to remove and replace spatial components, but the resultant HR-MS images show severe spectral distortion. The MRA-based methods inject high-frequency features of PAN derived by multi-resolution decomposition techniques into upsampled multi-spectral images. Decimated wavelet transform (DWT) \cite{DWT1989}, high-pass filter fusion (HPF) \cite{HPF}, Laplacian pyramid (LP) \cite{vivone2014critical}, smoothing filter-based intensity modulation (SFIM) \cite{SFIM}, and atrous wavelet transform (ATWT) \cite{ATWT1999} are typical MRA-based methods that reduce spectral distortion and improve resolution, but they  heavily rely on multi-resolution techniques, which may cause local spatial artifacts. In recent years,  VO-based methods are used because of the fine fusion effect on ill-posed problems. These various constraints can only inadequately reflect the limited structural relations of the images.

\noindent
\textbf{Deep Learning-based Methods.}
Deep learning-based methods have been widely used for pan-sharpening \cite{zhou2022mutual, yanmemory,pmlr-v202-zhou23f,10167672,10119207,10106462,10059132,NEURIPS2022_91a23b3e,10.1007/978-3-031-19797-0_16,10.1145/3503161.3547924,10.1145/3503161.3547774,9858176,zhou2023memory}. PNN \cite{masi2016pansharpening} uses three convolutional units to directly map the relationship between PAN, LR-MS, and HR-MS images. Inspired by PNN, a large number of pan-sharpening studies based on deep learning emerge. 
For example, PANNet \cite{yang2017pannet} adopts the residual learning module in Resnet \cite{resnet}. MSDCNN \cite{yuan2018multiscale} adds multi-scale modules on the basis of residual connection. SRPPNN \cite{cai2020super} refers to the design idea of SRCNN \cite{dong2015image}.  

Recently, some model-driven deep models with physical meaning emerge. The basic idea is to use prior knowledge to formulate optimization problems, then unfold the optimization algorithms and replace the steps in the algorithm with deep neural networks. For example, Xu \emph{et al.} \cite{Xu_2021_CVPR} propose the model-based deep learning network MHNet and GPPNN for pan-sharpening, respectively. In terms of the function design $\rm prox_{\lambda, \Omega}$ that takes for the embedded prior term, existing works  heuristically employ diverse network architectures in the black-box fashion, thus resulting in weak physical meanings. It motivates us to explore the task-driven customized prior with clear physical patterns.

\section{Methodology}
\subsection{Model Formulation}
In general, pan-sharpening aims to obtain the HR-MS image $\mathbf{H}$ from its degradation observation $\mathbf{L} =(\mathbf{H} \otimes \mathbf{K}) \downarrow_{s} + \mathbf{n_s}$,
where $\mathbf{K}$ and $\downarrow_{s}$ denote blur kernel and down-sampling operation, and $\mathbf{n_s}$ is usually assumed to be additive white Gaussian noise (AWGN). The degradation process by using the maximum a  posterior (MAP) principle can be reformulated as:
\begin{align}\label{eq:model}
     \min\limits_{\mathbf{H}}\frac{1}{2}\left|\left|\mathbf{L}-\mathbf{D}\mathbf{K}\mathbf{H}\right|\right|_2^2+\lambda\Omega(\mathbf{H}|\mathbf{P}),
\end{align}
where $\lambda$ is a hyper-parameter to weight the first term (data fidelity term accords with degradation) and the regularization term $\Omega(.)$. We solve the optimization problem using the half-quadratic splitting (HQS) algorithm. By introducing one auxiliary variables $\mathbf{U}$, Eq.~\eqref{eq:model} can be reformulated as a non-constrained optimization problem:
\begin{align}
\mathop{\min}\limits_{\mathbf{H},\mathbf{U}} 
    &\frac{1}{2}\left|\left|\mathbf{L}-\mathbf{D}\mathbf{K}\mathbf{H}\right|\right|_2^2+
    \frac{\eta}{2} \left|\left|\mathbf{U}-\mathbf{H}\right|\right|_2^2+\lambda\Omega(\mathbf{U}|\mathbf{P}), \label{eq:min_opt3}
\end{align}
where $\eta$ is the penalty parameter. When $\eta$ approaches infinity, Eq. \eqref{eq:min_opt3} converges to Eq. \eqref{eq:model}. Minimizing Eq. \eqref{eq:min_opt3} involves updating $\mathbf{U}$ and $\mathbf{H}$ alternately.

\noindent
\textbf{Updating $\mathbf{U}$.} Given the estimated HR-MS image $\mathbf{H}^{(k)}$ at iteration $k$, the auxiliary variable $\mathbf{U}$ can be updated as:
\begin{align}
    \mathbf{U}^{(k)}  = \mathop{\arg\min}\limits_{\mathbf{U}} \frac{\eta}{2}\left|\left|\mathbf{U}-\mathbf{H}^{(k)}\right|\right|_2^2+\lambda\Omega(\mathbf{U}|\mathbf{P}).   \label{eq:opt_u11}
\end{align}
We can derive the solution of  Eq. (\ref{eq:opt_u11}) as
\begin{align}
    \rm \mathbf{U}^{(k)}=  {\rm prox}_{\Omega(.),\lambda,\eta}(\mathbf{H}^{(k)}, P),      \label{uk1}
\end{align}
where ${\rm prox}_{\Omega}(\cdot)$ is the proximal operator corresponding to the implicit local prior $\Omega(\cdot)$.

\noindent
\textbf{Updating $\mathbf{H}$.} Given $\mathbf{U}^{(k)}$, $\mathbf{H}$ is updated as:
\begin{align}
    \mathbf{H}^{(k+1)}  &=\mathop{\arg\min}\limits_{\mathbf{H}} 
    \frac{1}{2}\left|\left|\mathbf{L}-\mathbf{D}\mathbf{K}\mathbf{H}\right|\right|_2^2 
      +\frac{\eta}{2}\left|\left|\mathbf{U}^{(k)}-\mathbf{H}\right|\right|_2^2. \label{eq:opt_h1}
\end{align}
By applying the proximal gradient method \cite{1976Monotone} to Eq. (\ref{eq:opt_h1}), we update $\mathbf{H}$ using the gradient decent method. Consequently, the updated equation for $\mathbf{H}$ is
\begin{align}
    \mathbf{H}^{(k+1)}= \mathbf{H}^{(k)}-\delta_2\nabla f_2(\mathbf{H}^{(k)}), \label{hk1}
\end{align}
where $\delta_2$ is the step size, and the gradient $\nabla f_2(\mathbf{H}^{(k)})$ is 
\begin{align}
    \nabla f_2(\mathbf{H}^{(k)}) &= (\mathbf{D}\mathbf{K})^T(\mathbf{D}\mathbf{K}\mathbf{H}^{(k)}-\mathbf{L})  + \eta(\mathbf{H}^{(k)}-\mathbf{U}^{(k)}),   \label{eq:iter_gh2}
\end{align}
where $T$ is the matrix transpose operation.

\subsection{Model Flowchart} 
As detailed in Figure \ref{mainfig}, the proposed image reasoning prior-embedded framework consists of four training stages: 
\textbf{Stage 1}: Image Reasoning Prior Pre-training; \textbf{Stage 2}: Spatial-Spectral Consistency Pre-training; \textbf{Stage 3}: Penetrating Reasoning Prior into Unfolding Network; 
\textbf{Stage 4}: Spatial-Spectral Consistency Constraint as loss function.

\begin{figure}[t]
	\centering
	\includegraphics[width=0.5\textwidth]{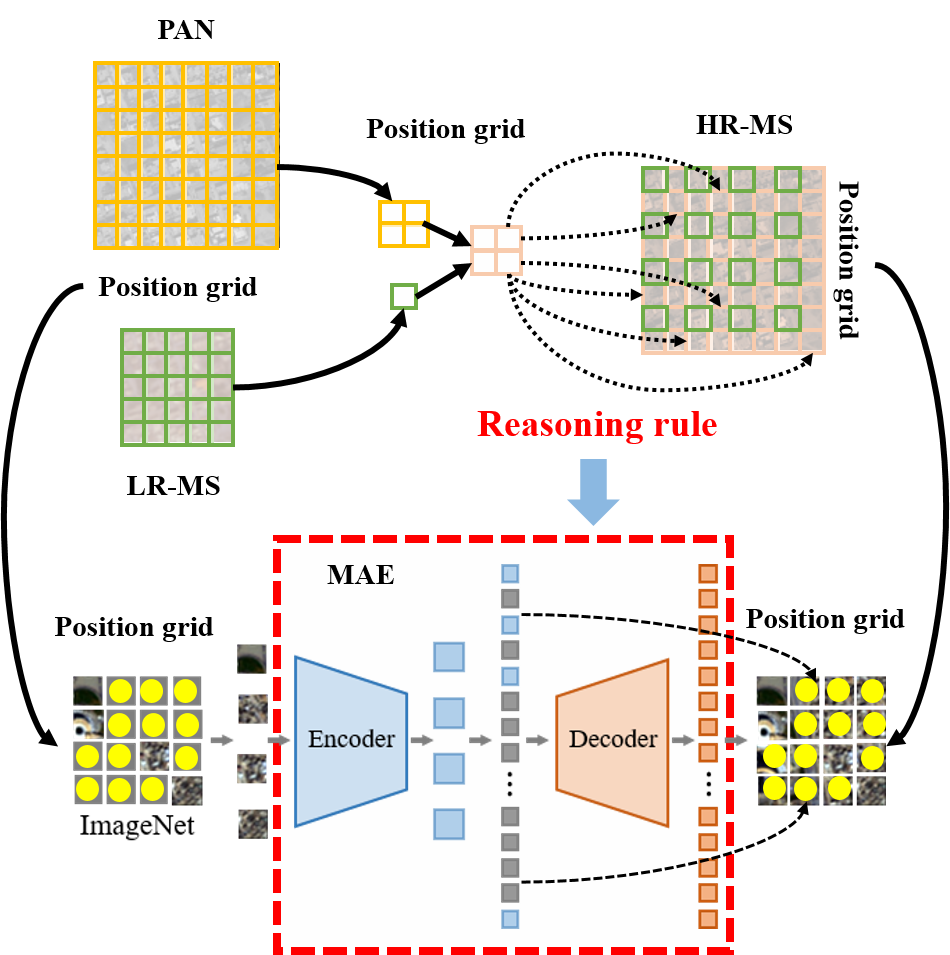}
	\caption{\textbf{Motivation.} Masked Autoencoders as the image reasoning learners. During pre-training, a large
random subset of image patches is masked out, remarked as yellow ones. The
encoder is applied to the small subset of visible patches. Mask tokens are introduced after the encoder, and the full set of encoded patches and mask tokens is processed by a decoder that reconstructs the original image in pixels. }
	\label{imagemae}
	\vspace{-3mm}
\end{figure}

\subsection{Structure Flow} \label{structure}
Based on the iterative algorithm, we construct deep unfolding network for pan-sharpening as shown in Figure~\ref{mainfig}. This network is an implementation of the algorithm for solving Eq.~(\ref{eq:model}). In terms of the function design $\rm prox_{\lambda, \Omega}$ that takes for the embedded prior term $\Omega(.)$, we stand on the shoulder of MAE proposed by He \emph{et al}, where the MAE is trained in a self-supervised manner and empowered with the reasoning ability, acting as the image prior. Based on the above analysis, the MAE is naturally treated as the prior term $\Omega(.)$ and meets the function of pan-sharpening.   
%


\textbf{Stage 1}: Based on the original MAE, we employ the pure convolution neural network as the encoder and decoder to implement its network architecture with masking patch strategies: 
(1) evenly divide the input image, randomly sample the small subset of regions and mask the remaining ones while keeping the whole image architecture; and
(2) both the small subset of visible patches and mask tokens are  processed by the encoder and decoder that reconstructs the original image in pixels. Note that the input into the encoder is the whole image, not the image patch.
%

  
  
  

\begin{figure}[t]
\begin{center}
\includegraphics[width=\columnwidth]{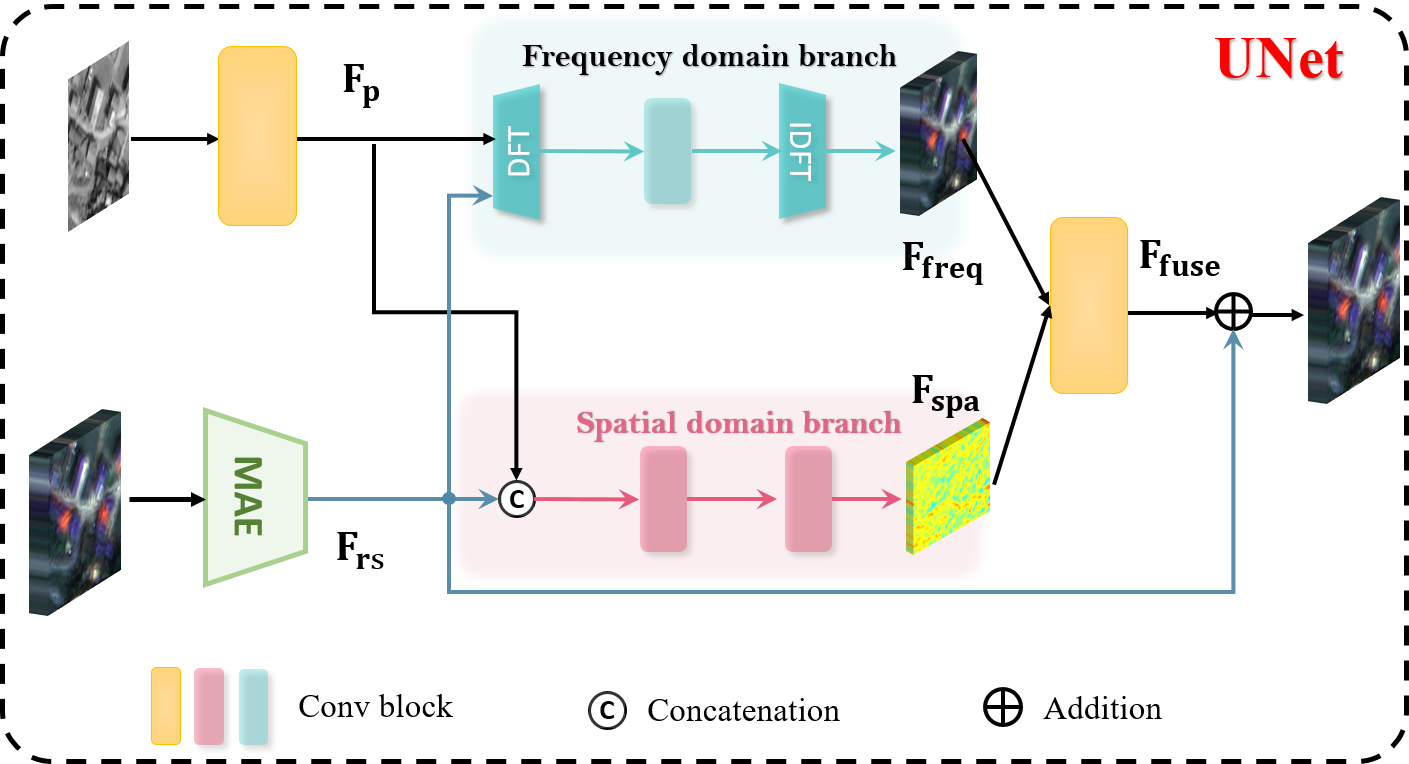}
\caption{The detailed structure of $\rm UNet$.}
\label{sonfig}
\end{center}
\vspace{-2em}
\end{figure}

\textbf{Stage 3}: \textbf{UNet.} Based on the pre-trained MAE encoder $\rm f_{CMAE}(.)$, we implement the whole architecture of $\mathbf{UNet}$ that is presented in Figure \ref{sonfig}. To be specific, the $\rm k$-th iteration $\mathbf{H}^{(k-1)}$ is fed into the encoder part $\rm E_{CMAE}(.)$ of the pre-trained MAE to generate the reasoning feature representation as
\begin{equation}
    \rm \mathbf{H}_{rs} = E_{CMAE}(\mathbf{H}^{(k-1)}),
\end{equation}
Then, PAN image $\rm P$ is projected into the shallow feature space by the convolution units as 
\begin{equation}
    \rm \mathbf{F}_{p} = Conv(\mathbf{P}).
\end{equation}
Referring to the representation $\rm \mathbf{H}_{rs}$ and the texture-rich PAN information $\rm \mathbf{F}_{p}$, we further incorporate them to reconstruct the HR-MS image by the spatial-frequency information transformation module $SFT$ that derived from \cite{zhou2022spatial}, which is  shown in Figure \ref{sonfig}. The information transformation process is detailed as
\begin{equation}
\rm F_{fuse} = SFT(H_{rsl}, F_{p}). 
\end{equation}

\textbf{HNet.} 
To transform the update process of $\mathbf{H}^{(k+1)}$ in Eq.~\eqref{hk1} into a network. Firstly, we need to implement two operations, i.e.,  $Down\downarrow_s$ and $Up\uparrow_s$, using the network. Specifically, $Down\downarrow_s$ is implemented by a spatial identify transformation convolution operator, and an additional $s$-strides followed convolution module with spatial resolution reduction: 
\begin{align}
\centering
    &\mathbf{DKH}^{(k)} = {\rm Conv}^{(s)}\downarrow(\mathbf{KH}^{(k)}), \label{hdown}
\end{align}
where ${\rm Conv}\downarrow_{(s)}$ aims to perform the $s$ times down-sampling. The latter operation $Up\uparrow_s$ is implemented by a deconvolution layer containing the $s$-strides convolution module with spatial resolution expansion and a convolution module with spatial identify transformation:
\begin{align}
\centering
    &\mathbf{UH}^{(k)} = {\rm Conv}^{(s)}\uparrow(\mathbf{L}-\mathbf{DKH}^{(k)}),  \label{hup}
\end{align}
where ${\rm Conv}\uparrow_{(s)}$ aims to perform the $s$ times up-sampling.

\begin{figure}[t]
	\centering
	\includegraphics[width=0.5\textwidth]{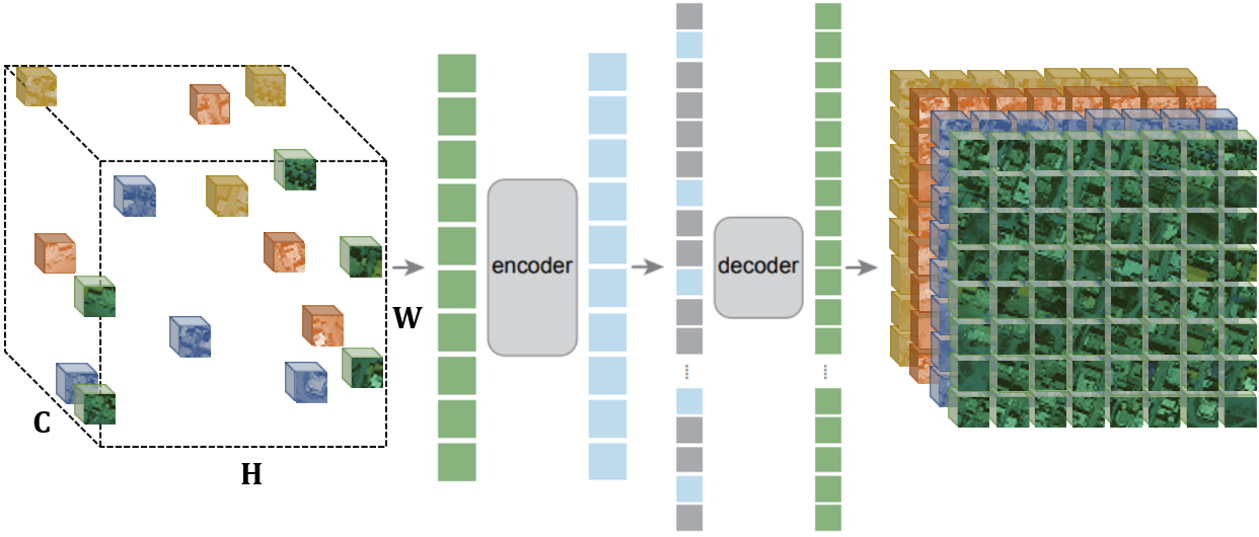}
	\caption{\textbf{Motivation.} Masked Autoencoders as spatiotemporal learners. It masks a large subset of random patches in spacetime. An encoder operates on a set of visible patches. A  decoder then processes the full set of encoded patches and mask tokens to reconstruct the input. Except for patch and positional embeddings,  the encoder, the decoder, and the masking strategy have no spatiotemporal inductive bias.}
	\label{videomae}
	\vspace{-3mm}
\end{figure}

\begin{table*}
\caption{Quantitative comparison with the state-of-the-art methods. The best results are highlighted in \textbf{bold}.}
\centering
\renewcommand{\tabcolsep}{3.2pt} 
\renewcommand{\arraystretch}{1.1} 
\resizebox{1.0\linewidth}{!}{\begin{tabular}{l|cccc|cccc|cccc}
\toprule 
\multirow{2}{*}{Method} & \multicolumn{4}{c|}{WordView II} & \multicolumn{4}{c|}{GaoFen2} & \multicolumn{4}{c}{WordView III} \\ \cline{2-13} 
 & PSNR $\uparrow$ & SSIM$\uparrow$ & SAM$\downarrow$ & ERGAS$\downarrow$ & PSNR $\uparrow$ & SSIM$\uparrow$ & SAM$\downarrow$ & ERGAS$\downarrow$ & PSNR $\uparrow$ & SSIM$\uparrow$ & SAM$\downarrow$ & ERGAS$\downarrow$ \\ \midrule
SFIM & 34.1297 & 0.8975 & 0.0439 & 2.3449 & 36.9060 & 0.8882 & 0.0318 & 1.7398 & 21.8212 & 0.5457 & 0.1208 & 8.9730 \\
Brovey & 35.8646 & 0.9216 & 0.0403 & 1.8238 & 37.7974 & 0.9026 & 0.0218 & 1.3720 & 22.506 & 0.5466 & 0.1159 & 8.2331 \\
GS & 35.6376 & 0.9176 & 0.0423 & 1.8774 & 37.2260 & 0.9034 & 0.0309 & 1.6736 & 22.5608 & 0.5470 & 0.1217 & 8.2433 \\
IHS & 35.2962 & 0.9027 & 0.0461 & 2.0278 & 38.1754 & 0.9100 & 0.0243 & 1.5336 & 22.5579 & 0.5354 & 0.1266 & 8.3616 \\
GFPCA & 34.5581 & 0.9038 & 0.0488 & 2.1411 & 37.9443 & 0.9204 & 0.0314 & 1.5604 & 22.3344 & 0.4826 & 0.1294 & 8.3964 \\\midrule
PNN (RS'16) & 40.7550 & 0.9624 & 0.0259 & 1.0646 & 43.1208 & 0.9704 & 0.0172 & 0.8528 & 29.9418 & 0.9121 & 0.0824 & 3.3206 \\
PANNet (ICCV'17) & 40.8176 & 0.9626 & 0.0257 & 1.0557 & 43.0659 & 0.9685 & 0.0178 & 0.8577 & 29.684 & 0.9072 & 0.0851 & 3.4263 \\
MSDCNN (TGRS'21) & 41.3355 & 0.9664 & 0.0242 & 0.9940 & 45.6874 & 0.9827 & 0.0135 & 0.6389 & 30.3038 & 0.9184 & 0.0782 & 3.1884 \\
SRPPNN (TGRS'20) & 41.4538 & 0.9679 & 0.0233 & 0.9899 & 47.1998 & 0.9877 & 0.0106 & 0.5586 & 30.4346 & 0.9202 & 0.0770 & 3.1553 \\
GPPNN (CVPR'21) & 41.1622 & 0.9684 & 0.0244 & 1.0315 & 44.2145 & 0.9815 & 0.0137 & 0.7361 & 30.1785 & 0.9175 & 0.0776 & 3.2596 \\

MutNet (CVPR'22) & 41.6773  & 0.9705 & 0.0224 & 0.9519 & 47.3042 & 0.9892 & 0.0102 & 0.5481 & 30.4907 & 0.9223 & 0.0749 & 3.1125 \\

MANet (ECCV'22) & 41.8577 & 0.9697 & 0.0229 & 0.9420 & 47.2668 & 0.9890 & 0.0102 & 0.5472 & 30.5451 & 0.9214 & 0.0769 & 3.1032 \\

\midrule
Ours&\textcolor{black}{\textbf{41.8735}}&\textcolor{black}{\textbf{ 0.9731 }}& \textcolor{black}{\textbf{0.0220}}&\textcolor{black}{\textbf{0.9413}}

& \textcolor{black}{\textbf{47.3931}} & \textcolor{black}{\textbf{0.9892}} & \textcolor{black}{\textbf{0.0089}} & \textcolor{black}{\textbf{0.5435}} 

& \textcolor{black}{\textbf{30.5560}} & \textcolor{black}{\textbf{0.9225}}&\textcolor{black}{\textbf{0.0733}}& \textcolor{black}{\textbf{3.0072}}\\
\bottomrule
\end{tabular}}
\label{taball}
\end{table*}

\begin{figure*}[h]
\begin{center}
\includegraphics[width=0.98\textwidth]{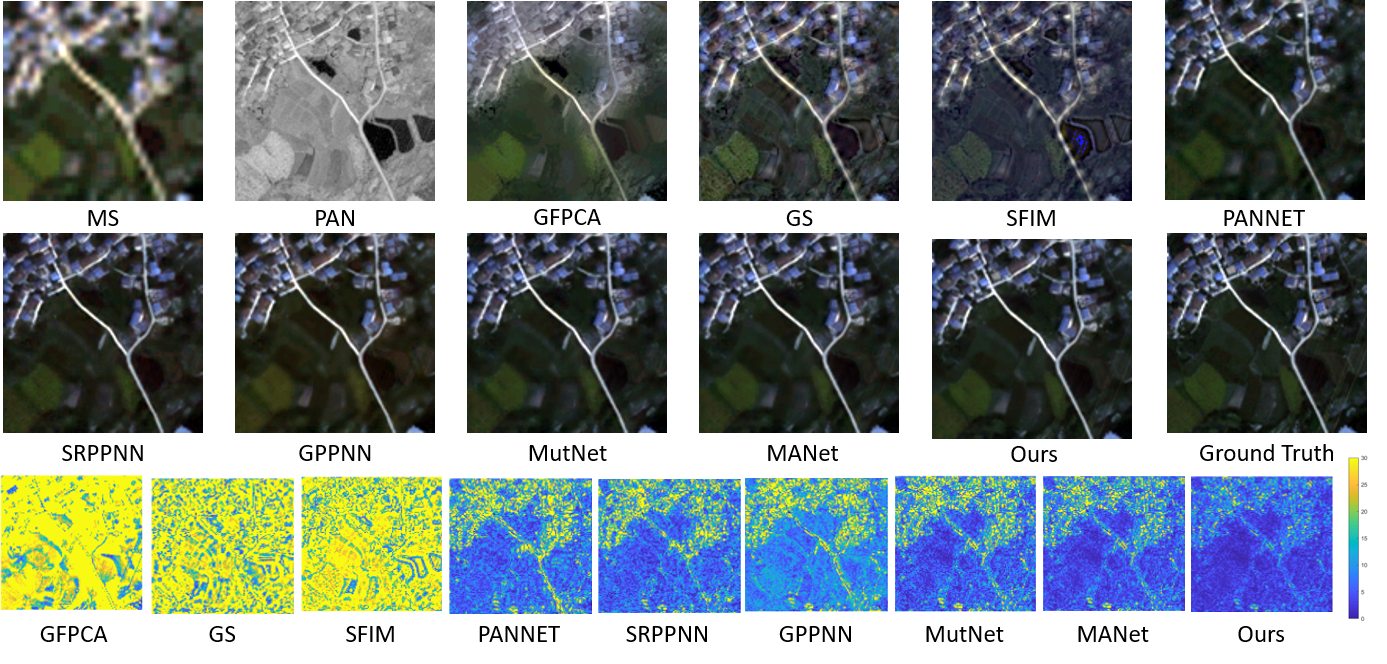}
\vspace{-14pt}
\caption{Visual comparison of the  HR-MS images produced by different methods for processing the LR-MS image from the WorldView-II dataset. Images  in  the  last  row visualize the mean squared error image between the output and ground truth.}
\label{wv2}
\end{center}
\vspace{-1.4em}
\end{figure*}

\begin{table*}
\normalsize
\centering
\renewcommand\arraystretch{1.2}
\caption{The average quantitative results on the GaoFen2 dataset in the full resolution case.}
\resizebox{1.0\linewidth}{!}{
\begin{tabular}{cccccccccccccc}
\toprule
 Metrics& SFIM& GS& Brovey& IHS & GFPCA & PNN & PANNET & MSDCNN & SRPPNN & GPPNN & MutNet & MANet& \textbf{Ours} \\ \midrule
$D_\lambda\downarrow$ & 0.0822 & 0.0696 & 0.1378& 0.0770 & 0.0914 & 0.0746& 0.0737 & 0.0734 & 0.0767 & 0.0782 & 0.0694 & 0.0681 & \textbf{0.0676} \\
$D_s\downarrow$ & \textbf{0.1087}& 0.2456 & 0.2605 & 0.2985& 0.1635& 0.1164 & 0.1224 & 0.1151 & 0.1162 & 0.1253 & 0.1118 & 0.1119 & 0.1112 \\
$QNR\uparrow$& 0.8214 & 0.7025& 0.6390& 0.6485& 0.7615& 0.8191& 0.8143 & 0.8251 & 0.8173 & 0.8073 & 0.8259 & 0.8266 & \textbf{0.8287} \\
\bottomrule
\end{tabular}
}
\label{tab-full}
\end{table*}

\begin{figure*}[h]
	\centering
	\includegraphics[width=0.98\textwidth]{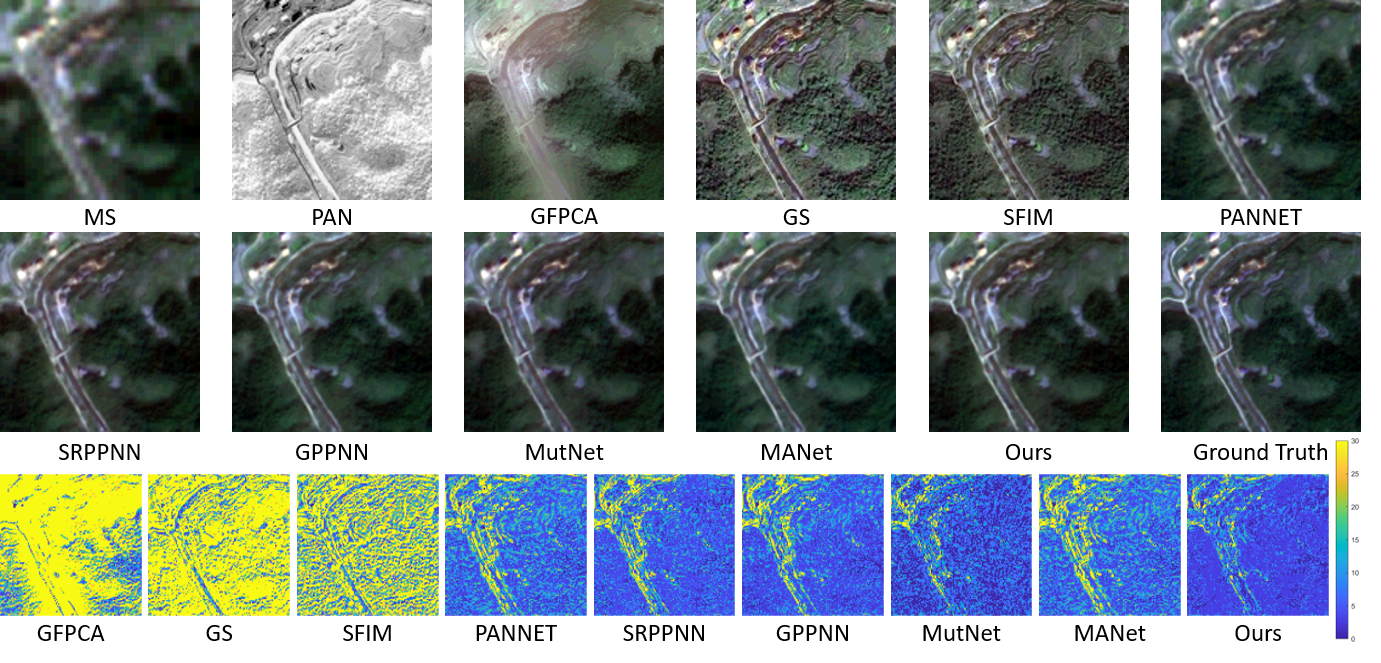} 
	\caption{Qualitative visualization comparison of our method with other representative counterparts on a typical satellite image pair from the GaoFen2 dataset. Images in the last row visualizes the MSE between the output and  ground truth.}
	\label{f3}
\end{figure*}

\subsection{Optimization Flow} \label{DUN}

\textbf{Stage 2}: To highlight, the uniqueness of our proposed method is that the entire learning process is fully and explicitly integrated with the inherent physical mechanism underlying the pan-sharpening task. Specifically, based on the MAE with spatial-spectral masking strategy that is tailored with the spatial-frequency representation learner, we redevelop the MAE as the regularization term within the loss function to constrain the spatial-spectral consistency of the model output and its corresponding ground truth. As shown in Figure \ref{videomae}, standing on the shoulders of the video-version extension of masked Autoencoders \cite{feichtenhofer2022masked} proposed by He \emph{et al}, we redevelop the masked image modeling as ``learned loss function'' to constraint the spatial-spectral representation consistency by the following implementation details:  

\begin{itemize}
  \item  randomly mask out spatial-spectral patches in the ground truth image and then learn an autoencoder to reconstruct them;

  \item the only spatial-spectral specific inductive bias is on embedding the patches and their positions; all other components are agnostic to the spatial-spectral nature of the problem. In particular, the encoder and decoder are both vanilla Vision Transformers with no factorization or hierarchy, and our random mask sampling is agnostic to the spatial-spectral structures.
\end{itemize}

To generate pleasing pan-sharpening results, we construct our training objective function using the mean absolute error loss over image-level measurement, which is defined as
\begin{equation}\label{LossFunction}
 \mathbf{L_{img}} =\sum_{i=1}^{N}\left \| \mathbf{H}_{i}^{(K+1)} - \mathbf{H}_{gt,i}  \right \|_{1},
\end{equation}
where $\mathbf{H}_{i}^{(K+1)}$ denotes the $i$-th estimated HR-MS image, $\mathbf{H}_{gt,i}$ is $i$-th ground truth HR-MS image, and $N$ is the number of training pairs.

\textbf{Stage 4}: Suppose that the pre-trained MAE model is $f_{mae}(.)$ and its encoder part is $E_{mae}(.)$, it is employed as the complementary loss function to the original image-level loss function $\mathbf{L_{img}}$ as 
\begin{equation}
    \mathbf{L_{ss}} = \sum_{i=1}^{N}||E_{mae}(\mathbf{H}_{gt,i})-E_{mae}(\mathbf{H}_{i}^{(K+1)})||_{1}.
\end{equation}
The total loss function is remarked as 
\begin{equation}
    \mathbf{L}= L_{img}+\lambda \times L_{ss}.
\end{equation}
where $\lambda$ is weighted factor and set as 1 in our work.

\section{Experiments}

\subsection{Settings}
\label{sec:dataset1}

\noindent
\textbf{Datasets.} 
Due to the unavailability of ground-truth MS images, we follow the previous works to generate the training set by employing the Wald protocol tool \cite{gt}. Specifically, given the MS image $\rm H\in R^{M \times N \times C}$ and the PAN image $\rm P_H \in R^{rM \times rN \times b}$, both of them are downsampled by a ratio $\rm r$, and then are denoted as $\rm L \in R^{M/r \times N/r \times C}$ and $\rm P \in R^{M \times N \times b}$, respectively. In the training set, $\rm L$ and $\rm p$ are regarded as the inputs, while $\rm H$ is the ground truth. In our work, three satellite images of  WorldView II, GaoFen2, and  WorldView III are adopted to construct image datasets. For each database, PAN images are cropped into patches with a size of $128 \times 128$ while the corresponding MS patches are with a size of $32 \times 32$.

\noindent
\textbf{Baselines.}
Several state-of-the-art Pan-sharpening methods are compared, including   seven representative deep learning-based methods: PNN \cite{masi2016pansharpening}, PANNET \cite{yang2017pannet}, MSDCNN \cite{yuan2018multiscale}, SRPPNN \cite{cai2020super}, GPPNN \cite{Xu_2021_CVPR}, MANet \cite{yanmemory}, and MutNet \cite{zhou2022mutual} and five promising traditional methods: SFIM \cite{SFIM}, Brovey \cite{Brovey}, GS \cite{GS}, IHS \cite{IHS}, and GFPCA \cite{GFPCA}.  

\noindent
\textbf{Metrics.}
Several widely-used image quality assessment (IQA) metrics are employed for performance measurement, including  PSNR, SSIM, SAM \cite{sam}, ERGAS \cite{ergas}, and  three non-reference metrics  $D_\lambda$,  $D_S$, and QNR for real-world full-resolution scenes. 

\noindent
\textbf{Implementations.}
In our experiments, all the designed networks are implemented with PyTorch framework and trained on the PC with a single NVIDIA GeForce GTX 3090 GPU. In the training phase, these networks are optimized by the Adam optimizer~\cite{kingma2017adam} over 1000 epochs with a mini-batch size of 4. The learning rate is initialized  with $5 \times 10^{-4}$. When reaching 200 epochs, the learning rate is decayed by multiplying 0.5.

\subsection{Comparisons}

\noindent
\textbf{Evaluation on reduced-resolution scenes.} The comparison results on three satellite datasets are reported in Table \ref{taball}. As can be seen, our proposed method achieves the best overall results than other pan-sharpening methods across all the satellite datasets. Specifically, the average gains of our method over the second-best MANet are 0.12dB, 0.32dB, and 0.10dB in terms of PSNR on WorldView-II, GaoFen2, and  WorldView-III datasets, respectively. In addition to PSNR, consistent improvements can be observed in the other metrics, indicating  lower spectral distortion and spatial texture preservation. Our method outperforms other compared methods by a large margin.  The corresponding visual comparisons shown in Figure \ref{wv2} and Figure \ref{f3} also support the above claim. More visual results can be found in the supplementary material.

\noindent
\textbf{Evaluation on full-resolution scenes.} In order to demonstrate the real-world application, we further perform experiments on 200 sets of full-resolution data obtained by the additional Gaofen2. Due to the unavailability of ground-truth MS images in  real-world full-resolution scenes, the commonly-used three non-reference metrics of $D_\lambda$, $D_s$, and QNR are adopted for evaluation. The quantitative comparisons between representative deep learning-based methods and our method are shown in Table \ref{tab-full}. Our methods surpass other pan-sharpening methods in all metrics.

\begin{table}[h]
\centering
\renewcommand{\arraystretch}{1.1}
\caption{Quantitative results of the model with different number stages.}
\resizebox{1.0\linewidth}{!}{\begin{tabular}{c|cccc}
\toprule
Stage Number (K)& PSNR$\uparrow$ & SSIM$\uparrow$&SAM$\downarrow$&ERGAS$\downarrow$ \\ \midrule
1 &  41.2459 & 0.9655  &    0.0250  &  1.0123   \\
2  & 41.4962 &	0.9679 &	0.0240  &	0.9838  \\ 
3  & 41.7152 &	0.9722 &	0.0223  &	0.9506  \\ 
4  & \textbf{41.8735}	& \textbf{0.9731}	 & \textbf{0.0220}	& \textbf{0.9413}\\
5  & 41.8461 &	0.9697 &	0.0226 &	0.9421 \\
6 &41.7429 & 0.9733 &	0.0221 &	0.9506 \\
\bottomrule
\end{tabular}}
\label{tab1}
\end{table}

\begin{table}[h]
 \centering
 \caption{Quantitative results with the ablation of key components.}
 \normalsize
 \renewcommand{\tabcolsep}{3pt} 
 \renewcommand{\arraystretch}{1.0}
\resizebox{1.0\linewidth}{!}{
\begin{tabular}{cc|ccccccccc}
\toprule
$\rm U_{MAE}$ & $\rm L_{MAE}$ & PSNR$\uparrow$ & SSIM$\uparrow$ &SAM$\downarrow$ & ERGAS$\downarrow$ \\
 \midrule
  &   &    41.4655&0.9669 & 0.0253 &0.9724  \\
\checkmark &    &41.6576 	&0.9681 	&0.0241 	&0.9679  \\
 &  \checkmark &    41.8382 &0.9695 & 0.0231 &0.9423  \\

\checkmark  & \checkmark   &\textbf{41.8735} 	&\textbf{0.9731} &\textbf{0.0220} &\textbf{0.9413} \\

\bottomrule
\end{tabular}
}
\label{tab2}
\end{table}

\subsection{Ablation Study}
 To explore the contribution of different hyper-parameters and the key components, we conduct  ablation studies on the WorldView-II dataset.

\noindent
\textbf{Impact of the Stage Numbers.} To investigate the impact of the number of unfolded stages, we experiment proposed method with varying numbers of stages $\rm K$. Observing the results from Table \ref{tab1}, we found that the model's performance has obtained considerable improvement as the number of stages increases until reaching  4. When further increasing the $\rm K$, the results show a decreasing trend, which may be caused by the difficulty of gradient propagation. We set $\rm K=4$ as default stage number to balance the performance and computational complexity.

\noindent
\textbf{Effect of Key Components.} To investigate the contribution of the devised modules in our network, we take the model with $\rm K=4$ as the baseline and then conduct the comparison by observing the difference before and after removing the components. The corresponding  quantitative comparisons are reported in Table \ref{tab2}, where $\rm U_{MAE}$ represents the MAE within the $\rm UNet$ network  and $\rm L_{MAE}$ represents the MAE within loss function. As can be observed, equipping with both the MAE priors  significantly improves the model performance. 

\begin{figure}[h]
\begin{center}
\includegraphics[width=\columnwidth]{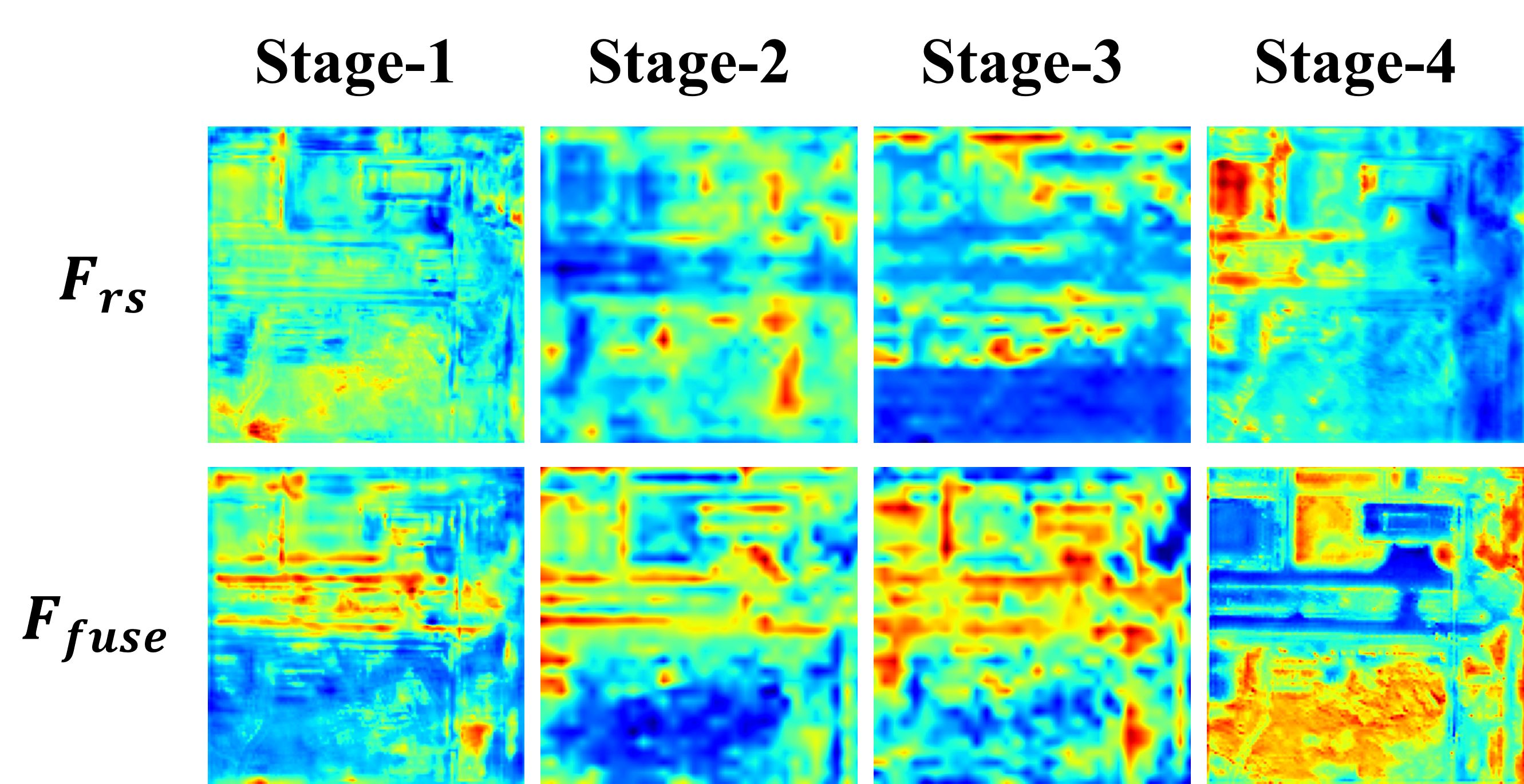}
\caption{The feature visualization in $\rm Unet$ upon the iterative stage increasing of deep unfolding network.}
\label{vis}
\end{center}
\vspace{-1em}
\end{figure}

 \subsection{Effect of MAE prior}
To verify the effect of the designed MAE prior, we deepen into the feature maps of  $F_{rs}$, $F_{fuse}$. As illustrated in section \ref{structure} that the MAE prior takes account for predicting the missing information of $F_{rs}$ and then enhances the representation $F_{fuse}$, with the stage increasing, the MAE-prior reasoned feature $F_{rs}$ is gradually enhanced and the resulted $F_{fuse}$ becomes more informative, thus supporting the powerful capability of the MAE prior, detailed in Figure \ref{vis}.



\section{Conclusion}
In this paper, we proposed the first work to focus on the designs of the deep prior term. We employ the learned MAE in a self-supervised manner acting as an image prior and then embed the pre-trained MAE with reasoning ability to penetrate deep unfolding architecture, thus making it more  transparent.   We also redevelop the pre-trained MAE with a spatial-spectral masking strategy and employ it as the regularization term within loss function to constrain the spatial-spectral consistency. The contained intrinsic knowledge over MAE loss term empowers the main unfolding network learning ability. Extensive experiments on three satellite datasets demonstrate its superiority.

{\small
\bibliographystyle{ieee_fullname}
\bibliography{main}

\begin{thebibliography}{10}\itemsep=-1pt

\bibitem{ergas}
L. Alparone, L. Wald, J. Chanussot, C. Thomas, P. Gamba, and L.~M. Bruce.
\newblock Comparison of pansharpening algorithms: Outcome of the 2006 grs-s
  data fusion contest.
\newblock {\em IEEE Transactions on Geoscience and Remote Sensing},
  45(10):3012--3021, 2007.

\bibitem{cai2020super}
Jiajun Cai and Bo Huang.
\newblock Super-resolution-guided progressive pansharpening based on a deep
  convolutional neural network.
\newblock {\em IEEE Transactions on Geoscience and Remote Sensing}, 2020.

\bibitem{carper1990use}
WJOSEPH CARPER, THOMASM LILLESAND, and RALPHW KIEFER.
\newblock The use of intensity-hue-saturation transformations for merging spot
  panchromatic and multispectral image data.
\newblock {\em Photogrammetric Engineering and remote sensing}, 56(4):459--467,
  1990.

\bibitem{dong2015image}
Chao Dong, Chen~Change Loy, Kaiming He, and Xiaoou Tang.
\newblock Image super-resolution using deep convolutional networks.
\newblock {\em IEEE Transactions on Pattern Analysis and Machine Intelligence},
  38(2):295--307, 2015.

\bibitem{feichtenhofer2022masked}
Christoph Feichtenhofer, Haoqi Fan, Yanghao Li, and Kaiming He.
\newblock Masked autoencoders as spatiotemporal learners.
\newblock {\em arXiv preprint arXiv:2205.09113}, 2022.

\bibitem{gillespie1987color}
Alan~R Gillespie, Anne~B Kahle, and Richard~E Walker.
\newblock Color enhancement of highly correlated images. ii. channel ratio and
  "chromaticity" transformation techniques.
\newblock {\em Remote Sensing of Environment}, 22(3):343--365, 1987.

\bibitem{Brovey}
A.~R. Gillespie, A.~B. Kahle, and R.~E. Walker.
\newblock Color enhancement of highly correlated images. ii. channel ratio and
  "chromaticity" transformation techniques - sciencedirect.
\newblock {\em Remote Sensing of Environment}, 22(3):343--365, 1987.

\bibitem{IHS}
R. Haydn, G.~W. Dalke, J. Henkel, and J.~E. Bare.
\newblock Application of the ihs color transform to the processing of
  multisensor data and image enhancement.
\newblock {\em National Academy of Sciences of the United States of America},
  79(13):571--577, 1982.

\bibitem{he2022masked}
Kaiming He, Xinlei Chen, Saining Xie, Yanghao Li, Piotr Doll{\'a}r, and Ross
  Girshick.
\newblock Masked autoencoders are scalable vision learners.
\newblock In {\em Proceedings of the IEEE/CVF Conference on Computer Vision and
  Pattern Recognition}, pages 16000--16009, 2022.

\bibitem{resnet}
Kaiming He, Xiangyu Zhang, Shaoqing Ren, and Jian Sun.
\newblock Deep residual learning for image recognition.
\newblock In {\em IEEE Conference on Computer Vision and Pattern Recognition},
  pages 770--778, 2016.

\bibitem{10119207}
Xuanhua He, Keyu Yan, Jie Zhang, Rui Li, Chengjun Xie, Man Zhou, and Danfeng
  Hong.
\newblock Multiscale dual-domain guidance network for pan-sharpening.
\newblock {\em IEEE Transactions on Geoscience and Remote Sensing}, 61:1--13,
  2023.

\bibitem{kingma2017adam}
Diederik~P. Kingma and Jimmy Ba.
\newblock Adam: A method for stochastic optimization, 2017.

\bibitem{kwarteng1989extracting}
P Kwarteng and A Chavez.
\newblock Extracting spectral contrast in landsat thematic mapper image data
  using selective principal component analysis.
\newblock {\em Photogrammetric Engineering and remote sensing}, 55(339-348):1,
  1989.

\bibitem{GS}
Craig~A Laben and Bernard~V Brower.
\newblock Process for enhancing the spatial resolution of multispectral imagery
  using pan-sharpening, 2000.
\newblock US Patent 6,011,875.

\bibitem{GFPCA}
W. Liao, H. Xin, F.~V. Coillie, G. Thoonen, and W. Philips.
\newblock Two-stage fusion of thermal hyperspectral and visible rgb image by
  pca and guided filter.
\newblock In {\em Workshop on Hyperspectral Image and Signal Processing:
  Evolution in Remote Sensing}, 2017.

\bibitem{SFIM}
J.~G. Liu.
\newblock Smoothing filter-based intensity modulation: A spectral preserve
  image fusion technique for improving spatial details.
\newblock {\em International Journal of Remote Sensing}, 21(18):3461--3472,
  2000.

\bibitem{DWT1989}
SG Mallat.
\newblock A theory for multiresolution signal decomposition: The wavelet
  representation.
\newblock {\em IEEE Transactions on Pattern Analysis and Machine Intelligence},
  11(7):674--693, 1989.

\bibitem{masi2016pansharpening}
Giuseppe Masi, Davide Cozzolino, Luisa Verdoliva, and Giuseppe Scarpa.
\newblock Pansharpening by convolutional neural networks.
\newblock {\em Remote Sensing}, 8(7):594, 2016.

\bibitem{ATWT1999}
Jorge Nunez, Xavier Otazu, Octavi Fors, Albert Prades, Vicenc Pala, and Roman
  Arbiol.
\newblock Multiresolution-based image fusion with additive wavelet
  decomposition.
\newblock {\em IEEE Transactions on Geoscience and Remote sensing},
  37(3):1204--1211, 1999.

\bibitem{1976Monotone}
R.~T. Rockafellar.
\newblock Monotone operators and the proximal point algorithm.
\newblock {\em Siam J Control Optim}, 14(5):877--898, 1976.

\bibitem{HPF}
Robert~A Schowengerdt.
\newblock Reconstruction of multispatial, multispectral image data using
  spatial frequency content.
\newblock {\em Photogrammetric Engineering and Remote Sensing},
  46(10):1325--1334, 1980.

\bibitem{shah2008efficient}
Vijay~P Shah, Nicolas~H Younan, and Roger~L King.
\newblock An efficient pan-sharpening method via a combined adaptive pca
  approach and contourlets.
\newblock {\em IEEE Transactions on Geoscience and Remote Sensing},
  46(5):1323--1335, 2008.

\bibitem{vivone2014critical}
Gemine Vivone, Luciano Alparone, Jocelyn Chanussot, Mauro Dalla~Mura, Andrea
  Garzelli, Giorgio~A Licciardi, Rocco Restaino, and Lucien Wald.
\newblock A critical comparison among pansharpening algorithms.
\newblock {\em IEEE Transactions on Geoscience and Remote Sensing},
  53(5):2565--2586, 2014.

\bibitem{gt}
Lucien Wald, Thierry Ranchin, and Marc Mangolini.
\newblock Fusion of satellite images of different spatial resolutions:
  Assessing the quality of resulting images.
\newblock {\em Photogrammetric Engineering and Remote Sensing}, 63:691--699, 11
  1997.

\bibitem{MHNet2019}
Qi Xie, Minghao Zhou, Qian Zhao, Deyu Meng, Wangmeng Zuo, and Zongben Xu.
\newblock Multispectral and hyperspectral image fusion by ms/hs fusion net.
\newblock In {\em CVPR}, pages 1585--1594, 2019.

\bibitem{Xu_2021_CVPR}
Shuang Xu, Jiangshe Zhang, Zixiang Zhao, Kai Sun, Junmin Liu, and Chunxia
  Zhang.
\newblock Deep gradient projection networks for pan-sharpening.
\newblock In {\em CVPR}, pages 1366--1375, June 2021.

\bibitem{yanmemory}
Keyu Yan, Man Zhou, Li Zhang, and Chengjun Xie.
\newblock Memory-augmented model-driven network for pansharpening.
\newblock 2022.

\bibitem{yang2017pannet}
Junfeng Yang, Xueyang Fu, Yuwen Hu, Yue Huang, Xinghao Ding, and John Paisley.
\newblock Pannet: A deep network architecture for pan-sharpening.
\newblock In {\em IEEE International Conference on Computer Vision}, pages
  5449--5457, 2017.

\bibitem{ye2019pan}
Fei Ye, Yecai Guo, and Peixian Zhuang.
\newblock Pan-sharpening via a gradient-based deep network prior.
\newblock {\em Signal Processing: Image Communication}, 74:322--331, 2019.

\bibitem{yuan2018multiscale}
Qiangqiang Yuan, Yancong Wei, Xiangchao Meng, Huanfeng Shen, and Liangpei
  Zhang.
\newblock A multiscale and multidepth convolutional neural network for remote
  sensing imagery pan-sharpening.
\newblock {\em IEEE Journal of Selected Topics in Applied Earth Observations
  and Remote Sensing}, 11(3):978--989, 2018.

\bibitem{sam}
Roberta~H Yuhas, Alexander F.~H Goetz, and Joe~W Boardman.
\newblock Discrimination among semi-arid landscape endmembers using the
  spectral angle mapper (sam) algorithm.
\newblock {\em Proc. Summaries Annu. JPL Airborne Geosci. Workshop}, pages
  147--149, 1992.

\bibitem{10106462}
Kaiwen Zheng, Jie Huang, Man Zhou, Danfeng Hong, and Feng Zhao.
\newblock Deep adaptive pansharpening via uncertainty-aware image fusion.
\newblock {\em IEEE Transactions on Geoscience and Remote Sensing}, 61:1--15,
  2023.

\bibitem{9858176}
Man Zhou, Jie Huang, Xueyang Fu, Feng Zhao, and Danfeng Hong.
\newblock Effective pan-sharpening by multiscale invertible neural network and
  heterogeneous task distilling.
\newblock {\em IEEE Transactions on Geoscience and Remote Sensing}, 60:1--14,
  2022.

\bibitem{pmlr-v202-zhou23f}
Man Zhou, Jie Huang, Chun-Le Guo, and Chongyi Li.
\newblock Fourmer: An efficient global modeling paradigm for image restoration.
\newblock In {\em Proceedings of the 40th International Conference on Machine
  Learning}, volume 202 of {\em Proceedings of Machine Learning Research},
  pages 42589--42601. PMLR, 23--29 Jul 2023.

\bibitem{10167672}
Man Zhou, Jie Huang, Danfeng Hong, Feng Zhao, Chongyi Li, and Jocelyn
  Chanussot.
\newblock Rethinking pan-sharpening in closed-loop regularization.
\newblock {\em IEEE Transactions on Neural Networks and Learning Systems},
  pages 1--15, 2023.

\bibitem{10.1145/3503161.3547924}
Man Zhou, Jie Huang, Chongyi Li, Hu Yu, Keyu Yan, Naishan Zheng, and Feng Zhao.
\newblock Adaptively learning low-high frequency information integration for
  pan-sharpening.
\newblock In {\em Proceedings of the 30th ACM International Conference on
  Multimedia}, MM '22, page 3375–3384, New York, NY, USA, 2022. Association
  for Computing Machinery.

\bibitem{10.1145/3503161.3547774}
Man Zhou, Jie Huang, Keyu Yan, Gang Yang, Aiping Liu, Chongyi Li, and Feng
  Zhao.
\newblock Normalization-based feature selection and restitution for
  pan-sharpening.
\newblock In {\em Proceedings of the 30th ACM International Conference on
  Multimedia}, MM '22, page 3365–3374, New York, NY, USA, 2022. Association
  for Computing Machinery.

\bibitem{zhou2022spatial}
Man Zhou, Jie Huang, Keyu Yan, Hu Yu, Xueyang Fu, Aiping Liu, Xian Wei, and
  Feng Zhao.
\newblock Spatial-frequency domain information integration for pan-sharpening.
\newblock In {\em Computer Vision--ECCV 2022: 17th European Conference, Tel
  Aviv, Israel, October 23--27, 2022, Proceedings, Part XVIII}, pages 274--291.
  Springer, 2022.

\bibitem{10.1007/978-3-031-19797-0_16}
Man Zhou, Jie Huang, Keyu Yan, Hu Yu, Xueyang Fu, Aiping Liu, Xian Wei, and
  Feng Zhao.
\newblock Spatial-frequency domain information integration for pan-sharpening.
\newblock In {\em Computer Vision -- ECCV 2022}, pages 274--291, Cham, 2022.
  Springer Nature Switzerland.

\bibitem{10059132}
Man Zhou, Keyu Yan, Xueyang Fu, Aiping Liu, and Chengjun Xie.
\newblock Pan-guided band-aware multi-spectral feature enhancement for
  pan-sharpening.
\newblock {\em IEEE Transactions on Computational Imaging}, 9:238--249, 2023.

\bibitem{zhou2022mutual}
Man Zhou, Keyu Yan, Jie Huang, Zihe Yang, Xueyang Fu, and Feng Zhao.
\newblock Mutual information-driven pan-sharpening.
\newblock In {\em Proceedings of the IEEE/CVF Conference on Computer Vision and
  Pattern Recognition}, pages 1798--1808, 2022.

\bibitem{zhou2023memory}
Man Zhou, Keyu Yan, Jinshan Pan, Wenqi Ren, Qi Xie, and Xiangyong Cao.
\newblock Memory-augmented deep unfolding network for guided image
  super-resolution.
\newblock {\em International Journal of Computer Vision}, 131(1):215--242,
  2023.

\bibitem{NEURIPS2022_91a23b3e}
man zhou, Hu Yu, Jie Huang, Feng Zhao, Jinwei Gu, Chen~Change Loy, Deyu Meng,
  and Chongyi Li.
\newblock Deep fourier up-sampling.
\newblock In {\em Advances in Neural Information Processing Systems},
  volume~35, pages 22995--23008. Curran Associates, Inc., 2022.

\bibitem{zhu2012sparse}
Xiao~Xiang Zhu and Richard Bamler.
\newblock A sparse image fusion algorithm with application to pan-sharpening.
\newblock {\em IEEE transactions on geoscience and remote sensing},
  51(5):2827--2836, 2012.

\end{thebibliography}
}

\end{document}